\pdfoutput=1

\documentclass[11pt]{article}

\usepackage[final]{acl}

\usepackage{times}
\usepackage{latexsym}
\usepackage{amsmath} 
\usepackage{tabularx}
\usepackage{multirow}
\usepackage{booktabs}
\usepackage{paralist}
\usepackage{color} 

\usepackage[T1]{fontenc}

\usepackage{microtype}
\usepackage{geometry} 
\usepackage{caption} 
\usepackage{array} 
\usepackage{rotating} 
\usepackage{svg}

\usepackage{xcolor}
\usepackage{inconsolata}
\usepackage{booktabs} 
\usepackage{amsfonts}
\usepackage{amssymb}
\allowdisplaybreaks
\usepackage{tabularray}
\usepackage{tabularx}
\usepackage{multirow}
\usepackage{arydshln}
\usepackage{CJKutf8}

\usepackage{stfloats}

\def\gD{{\mathcal{D}}}
\def\gT{{\mathcal{T}}}
\def\sR{{\mathbb{R}}}

\title{Sparse Adapter Fusion for Continual Learning in NLP}

\author{Min Zeng$^{1}$\thanks{~~Equal contributions.}, Xi Chen$^{1}$\footnotemark[1], Haiqin Yang$^{2}$\thanks{~~Corresponding authors.}, Yike Guo$^{1}$\footnotemark[2]\\
        \textsuperscript{1}Hong Kong University of Science and Technology \\ 
        \textsuperscript{2}Shenzhen Technology University \\ 
        \texttt{min.zeng.u@gmail.com}, \texttt{yanghaiqin@sztu.edu.cn}, \texttt{yikeguo@ust.hk}
        }

\begin{document}
\maketitle

\begin{abstract}
Continual learning in natural language processing plays a crucial role in adapting to evolving data and preventing catastrophic forgetting.  Despite significant progress, existing methods still face challenges, such as inefficient parameter reuse across tasks, risking catastrophic forgetting when tasks are dissimilar, and the unnecessary introduction of new parameters for each task, which hampers knowledge sharing among similar tasks.  To tackle these issues, we propose a {\bf Sparse Adapter Fusion Method (SAFM)}, which dynamically fuses old and new adapters to address these challenges.  SAFM operates in two stages: the decision stage and the tuning stage.  In the decision stage, SAFM determines whether to incorporate a new adapter, reuse an existing one, or add an empty adapter.  The architecture search procedure, designed to prioritize reusing or adding empty adapters, minimizes parameter consumption and maximizes reuse.  In the tuning stage, SAFM especially facilitates a layer-wise loss to encourage differentiation between adapters, effectively capturing knowledge within the same task.  Experimental results consistently show that SAFM outperforms state-of-the-art (SOTA) methods, achieving comparable performance while utilizing less than 60\% of the parameters\footnote{Code is available at \url{https://github.com/OzymandiasChen/SAFM}.}.
\end{abstract}

\section{Introduction}
Continual learning (CL) is a paradigm that emulates the human ability to learn and acquire knowledge~\citep {DBLP:journals/tmlr/MendezE23,DBLP:journals/pami/WangZSZ24} continuously.  It focuses on retaining previously learned information and transferring it effectively to master new tasks.  Learning continually is essential for models to adapt rapidly to evolving tasks.  As a result, continual learning methodologies have emerged, enabling models to seamlessly assimilate new information over time.  However, CL faces a significant challenge: catastrophic forgetting (CF), where a model's performance on earlier tasks deteriorates due to shifts in data distribution introduced by new tasks, potentially erasing previously acquired knowledge.

Recent approaches to mitigating CF can be categorized into three main types: regularization-based, rehearsal-based, and architectural-based methods~\citep{kirkpatrick2017overcoming,DBLP:conf/coling/BiesialskaBC20}.  Regularization-based methods~\cite{zenke2017continual,kirkpatrick2017overcoming,schwarz2018progress,aljundi2018memory} maintain performance on previous tasks by constraining updates to critical parameters.  However, excessive reliance on regularizers can overly restrict network parameters, limiting the model’s ability to learn new knowledge.  Rehearsal-based methods~\citep{lopez2017gradient,rebuffi2017icarl,sun2019lamol,mi2020continual} rely on storing past samples or generating pseudo-samples from earlier tasks, facing limitations due to memory constraints or the lack of authenticity in the generated pseudo-samples.  Architectural-based methods~\citep{madotto2021continual,zhang2022continual} mitigate CF by adding task-specific adapters to approximate each task~\citep{houlsby2019parameter}.  Notable publications, such as AdapterCL~\citep{madotto2021continual}, CPT4DST~\citep{zhu2022continual} and ACM~\citep{zhang2022continual}, have addressed CF through architectural modifications.  However, as the number of tasks grows, the model parameters increase linearly, highlighting the importance of learning shared information to reduce parameter redundancy.  While ACM aims to reduce parameters by reusing adapters from previous tasks, it still incurs high training costs as it does not reduce the adapters across tasks.

To tackle these challenges, we propose a {\bf Sparse Adapter Fusion Method (SAFM)} to reduce redundancy by eliminating unnecessary task-specific adapters while reusing those learned from earlier tasks.  This strategy enables the model to learn both local and global information simultaneously, improving performance and parameter efficiency.  In this context, local information refers to the unique data captured by adapters within the same task, while global information refers to task similarities that guide the reuse of adapters from previous tasks.  SAFM operates in two stages: the decision stage and the tuning stage.  In the decision stage, SAFM determines whether to add a new adapter, reuse an existing one, or add an empty adapter.  The architecture search procedure prioritizes reusing or adding empty adapters, thereby minimizing parameter consumption and optimizing reuse.  Once the adapter for the current task is determined, SAFM proceeds to the tuning stage, where it fine-tunes the model parameters using pseudo-replay and a layer-wise loss.  Concretely, this layer-wise loss, defined as the cosine similarity between adapters in adjacent layers, maximizes information conveyance by enhancing the distinction between adapters within the same task, further improving performance.

We highlight our key contributions as follows: 
\begin{compactitem}[\textbullet]
       \item We propose the {\bf Sparse Adapter Fusion Method (SAFM)}, a parameter-efficient continual learning method that mitigates CF by tending to reuse previously learned adapters or deploy empty adapters for new tasks. 
    \item SAFM introduces a layer-wise loss to promote the distinction adapters in adjacent layers within the same task, facilitating effective knowledge transfer and maximizing information conveyance with limited parameters.
    \item Experimental results demonstrate the superior performance of our SAFM over SOTA methods, achieving comparable performance while utilizing less than 60\% of the parameters employed by the SOTA models.  
\end{compactitem}

\section{Related Work}
Continual learning aims to acquire knowledge from new tasks while maintaining proficiency in previously learned tasks~\citep{DBLP:journals/tmlr/MendezE23,DBLP:journals/pami/WangZSZ24}.  Various approaches have been proposed to mitigate the issue of catastrophic forgetting, including regularization-based, rehearsal-based, and architectural-based methods.  These techniques have also been extended to apply in natural language processing, driving advancements in the field~\citep{DBLP:conf/coling/BiesialskaBC20}.

\textit{Regularization-based methods} add constraints to the important parameters of previous tasks.  For example, 
EWC~\citep{kirkpatrick2017overcoming} identifies crucial parameters and prevents substantial updates on them, thereby preserving performance on earlier tasks.  ARPER~\citep{mi2020continual} mitigates forgetting by combining prioritized exemplar replay with adaptive regularization inspired by EWC.
\textit{Rehearsal-based methods} mitigate forgetting by replaying real or pseudo-samples of previous tasks. For instance, 
LAMOL~\citep{sun2019lamol} utilizes a language model to generate pseudo-samples, thereby eliminating the need for additional memory storage to retain previous samples.  DCL~\citep{zengdirichlet} introduces a novel generative-based rehearsal method in CL, assuming a Dirichlet distribution on the latent variables instead of the original Gaussian one applied in Conditional Variational Autoencoders.  InsCL~\citep{wang-etal-2024-inscl} dynamically replays previous data by leveraging task similarity measured via the Wasserstein Distance.  It further prioritizes high-quality data using the Instruction Information metric (InsInfo), which evaluates instruction complexity and diversity.  PCGR \citep{chen2025prototype} proposes a Prototype Conditioned Generative
Replay (PCGR) method, which enhances generative reply by incorporating task-level statistics
through a Prototype Conditioned Variational Autoencoder (PCVAE).  \textit{Architectural-based approaches} reduce forgetting by modifying the network architecture.  For example, AdapterCL~\citep{madotto2021continual} places a residual adapter layer~\citep{houlsby2019parameter} atop each transformer layer to approximate different tasks.  More recently, CPT4DST~\citep{zhu2022continual} introduces prompt tuning to reduce forgetting in dialogue state tracking (DST), while ACM~\citep{zhang2022continual} reduces parameters by reusing modules from previous tasks.  However, ACM does not reduce the number of adapter layers per task, which can lead to parameter redundancy and result in the same high training cost.  
SAPT~\citep{zhao2024sapt}, on the other hand, employs a separate adapter for each new task and uses a shared attention framework to facilitate knowledge transfer between tasks. 
TCL~\citep{zeng2025task} reduces parameters by employing Task-wrapped Adapters (TWAs) to jointly learn both global and task-specific local information across tasks.

Overall, existing CL methods still face challenges related to parameter redundancy, resulting in high memory and computational costs.  Therefore, it is essential to develop approaches that reduce model parameters while preserving performance across tasks.

\section{Methodology}
\subsection{Task Definition}
\label{sec:def}

The goal of continual learning is to sequentially learn a set of tasks without catastrophically forgetting previously learned ones.  Formally, given a sequence of $N$ tasks ${\gT_1, \ldots, \gT_N}$ arriving in a streaming fashion, where each task $\gT_n$ consists of $N_n$ samples in $\gD_n=\{(x_n^i, y_n^i)\}_{i=1}^{N_n}$, CL aims to learn a function $f_\theta^n$ such that {the model must not only adapt to the current task $\gT_n$ but also maintain its performance on all previously learned tasks without forgetting.}

\begin{figure*}[!t]
  \centering
  \includegraphics[width=0.8\linewidth]{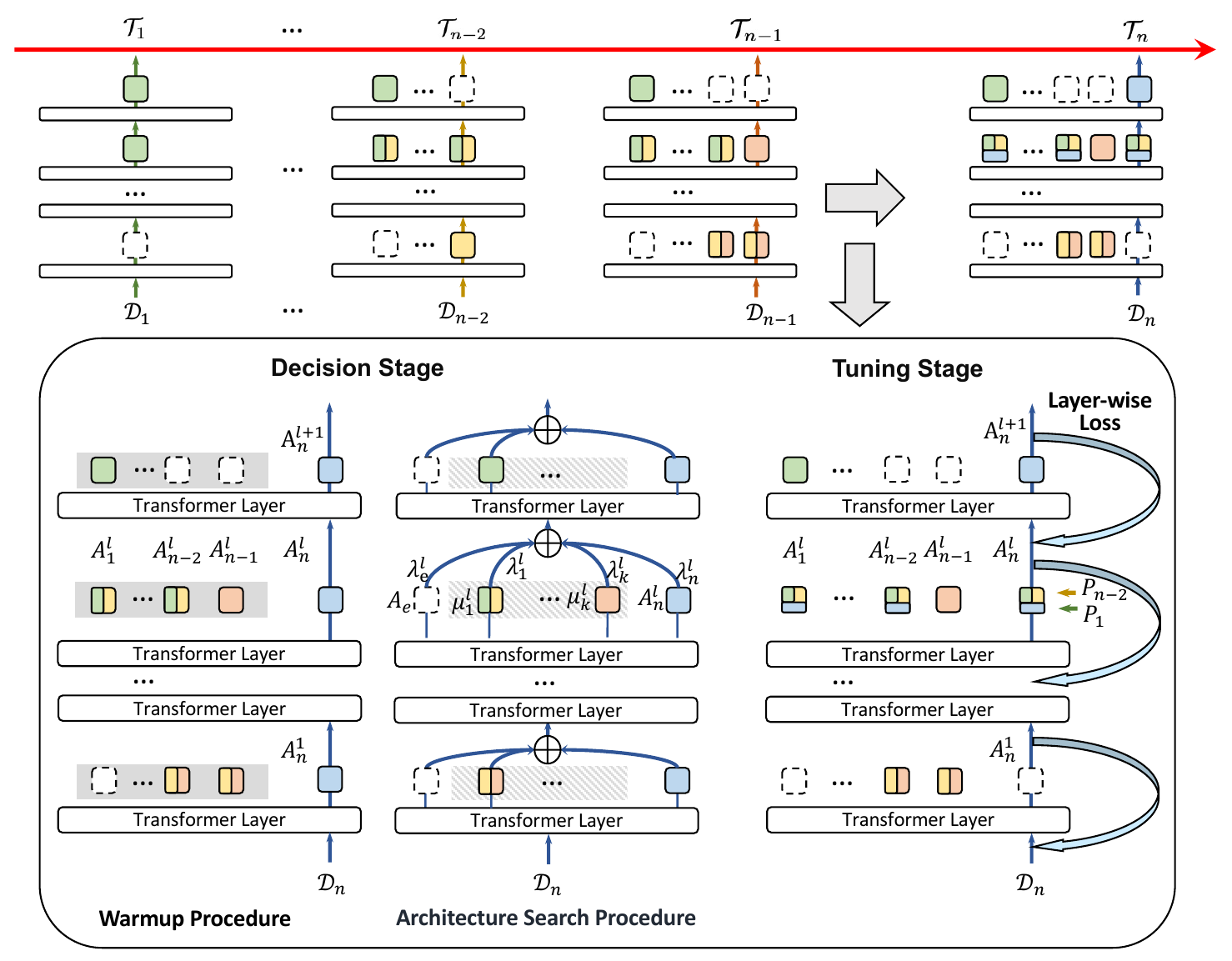}
  \caption{
  {SAFM consists of two stages: the decision stage and the tuning stage.  The color indicates a module is updated with data from a specific task: green for task $\gT_1$, yellow for task $\gT_{n-2}$, and blue for task $\gT_n$.  In the decision stage, a new module (blue) is initialized for task $\gT_n$, and the corresponding architecture search procedure is determined by Eq.~(\ref{eq:lambda}), which yields $A_n^l=A_1^l=A_{n-2}^l$.  Hence, at the tuning stage, $A_n^l$ has to be fine-tuned with data from task $\gT_1$ (green), $\gT_{n-2}$ (yellow), and $\gT_n$ (blue).  We then generate pseudo-samples from $\gT_1$ ($P_1$) and $\gT_{n-2}$ ($P_{n-2}$) with the incoming data in $\gD_n$ to update the module $A_n^l$.  For further details, please refer to Sec.~\ref{sec:archi}. 
  }
  }

\label{fig:Overall}
\vspace{-1em}
\end{figure*}

\subsection{Overview}
\label{sec:archi}
Figure~\ref{fig:Overall} illustrates the procedure of our proposed Sparse Adapter Fusion Method (SAFM), which consists of two stages: the decision stage and the tuning stage.  In the decision stage, when the task $\gT_n$ arrives, the SAFM works on top of each transformer layer.  Unlike ACM, which only decides whether to reuse an old adapter or add a new one for $\gT_n$, SAFM also considers the option of adding an empty adapter at each layer.  In other words, SAFM may choose not to insert any adapter or reuse the previous ones, which reduces the model's parameter size.  Once the architecture for task $\gT_n$ is determined, SAFM proceeds to the tuning stage, where it fine-tunes the model by using pseudo-replay to absorb the knowledge in previous tasks and a layer-wise loss to increase the distinction between adapters in adjacent layers within task $\gT_n$, which yields further performance improvement. 


\subsection{Decision Stage}
\label{sub:decision}
The decision stage consists of two critical procedures: the \textit{\textbf{warmup procedure}} and the \textit{\textbf{architecture search procedure}}.  When given task $\gT_n$, SAFM first enters the warmup procedure by initializing a task-specific adapter in each transformer layer, $A_n=\{A_n^1, \ldots, A_n^l, \ldots, A_n^L\}$, where $A_n^l$ represents the adapter at layer $l$ for task $\gT_n$, and $L$ is the total number of transformer layers (e.g., for GPT-2, $L=12$).  


{
Next, SAFM proceeds to the architecture search procedure layer-by-layer by the following steps:
\begin{compactenum}
    \item \textbf{Renaming Prior Adapters:} The unique adapters from previous tasks ${A_1^l, \ldots, A_{n-1}^l}$ at layer $l$ are relabeled as $\mu^l = \{\mu_1^l, \ldots, \mu_k^l\}$, where $k$ is the number of distinct adapters, constrained by $k \leq {t-1}$ due to the potential reuse of adapter layers.
    \item \textbf{Constructing Candidate Adapters:} Combine $\mu^l$ with the empty adapter $A_e$ and the newly initialized adapter $A_n^l$ to form $k+2$ candidate adapters: $c^l = \{A_e, \mu_1^l, \ldots, \mu_k^l, A_n^l\}$. 
    \item \textbf{Determining the final $A_n^l$:} The adapter from $c^l$ with the highest weight, computed by Eq.~(\ref{eq:lambda}), is selected as the final $A_n^l$.
\end{compactenum}
Let $h_n^l\in \sR^{d}$ be the hidden state of the adapter at layer $l$ for task $\gT_n$, where $d$ is the dimensionality of the embeddings and hidden states (For GPT-2, $d$ is 768).  $h_n^l$ can be expressed as a weighted average of the output hidden states of the candidate adapters:
\begin{align} \nonumber
h_n^l =& \lambda_{e}^l\!\times\! A_{e}(f_n^l(h_n^{l-1}))+ \sum_{i=1}^{k} \lambda_{i}^l\!\times\!\mu_{i}^l(f_n^l(h_n^{l-1})) \\ &+ \lambda_{n}^l \times A_{n}^l(f_n^l(h_n^{l-1})).\label{eq:hidden_state}
\end{align} 
Here, we slightly abuse the notation by using $A_{e}$, $\{\mu\}_{i=1}^l$, and $A_n^l$ to denote the parameters of the adapters.  Their weights are defined as $\lambda^l = \{\lambda_e^l, \lambda_1^l, \ldots, \lambda_k^l, \lambda_n^l\}$ accordingly, where $\lambda_e^l$ is the weight of the empty adapter, $\lambda_n^l$ is the weight of the newly initialized task-specific adapter for task $\gT_n$, and $\{\lambda_1^l, \ldots, \lambda_k^l\}$ represent the weights of the unique adapter modules from previous tasks.  $f_n^l$ denotes the $l$-th transformer layer of the language model for Task $\gT_n$.  
}

{
After the decision stage, we attain the parameters of Task $\gT_n$: $A_n=\{A_n^1, \ldots,  A_{n}^{l}, \ldots, A_n^L\}$.  For example, as shown in Fig.~\ref{fig:Overall}, $A_n^1=A_e$, $A_n^l=A_{n-2}^l=A_1^l$.  That is, the adapter at layer $1$ for task $\gT_n$ is empty, $A_e$.  The adapter at layer $l$ for task $\gT_n$ is the same as that for task $\gT_{n-2}$ and task $\gT_1$.  So $A_n$ contains at least an empty adapter and a reused adapter, which results in fewer model parameters. 


To learn the weight vector $\lambda^l\in\sR^{k+2}$, we define $\alpha\in\sR^+$ and $\beta\in\sR^+$ as the sparse factor and the reuse factor to determine the selection probabilities of $c^l$.  Then, $\lambda^l$ is initialized as a softmax function over $\{\alpha, \beta, \ldots, \beta, -\beta\}$, i.e., 
\begin{equation}\label{eq:lambda}
\lambda_{e}^l=\frac{e^{\alpha}}{\varphi},~~
\lambda_{j}^l
= \frac{e^{\beta}}{\varphi}, (j\in [1, k]),~~ \lambda_{n}^l =\frac{e^{-\beta}}{\varphi}
\end{equation}
where $\varphi = {e^{\alpha}+ k\times{e^{\beta}}+e^{-\beta}}$.  It is important to note that usually, we set $\alpha>\beta>0$ and yield a higher probability of selecting an empty adapter or reusing an existing adapter than creating a new one, as $\lambda_{e}^l=\lambda_{j}^le^{\alpha-\beta}>\lambda_{j}^l=\lambda_{n}^le^{2\beta}>\lambda_{n}^l$. 



}

\subsection{Tuning Stage}
\label{sub:training}
After the decision stage, SAFM determines the adapter architecture for task $\gT_n$ and processes to the tuning stage, where the adapter parameters are fine-tuned to better align with the training data distribution by applying the pseudo-replay mechanism and a layer-wise loss. 

For example, as illustrated in Fig.~\ref{fig:Overall}, suppose $A_{n}^l=A_{n-2}^l=A_1^l$, where $A_1^l$ is updated with data from task $\gT_1$ (green) and $\gT_{n-2}$ (yellow).  We then apply the \textbf{pseudo-replay} generation mechanism as ACM~\citep{zhang2022continual} to generate pseudo-samples from tasks $\gT_1$ and $\gT_{n-2}$, denoted as $P_{1}$ and $P_{n-2}$, respectively, and update the module $A_n^l$ with $P_{1}$, $P_{n-2}$, and incoming data $\gD_n$. 

After that, we place a \textbf{layer-wise loss} to enlarge the distance between adapters for each task, distinguishing the modules at each layer.  Specifically, the layer-wise loss between layer $l$ and layer $l-1$ of task $\gT_n$ is measured by the cosine similarity between the hidden state of two adjacent adapters:
\begin{equation}
\mathcal{L}_n^l=\left\{\begin{array}{ll}
   0  & \mbox{If } A_n^l=A_e \\
   \cos(h_n^l, h_n^{l-1})& \mbox{Otherwise} 
\end{array},
\right.
\end{equation}
where $h_n^l$ is computed by Eq.~(\ref{eq:hidden_state}) from $h_n^{l-1}$.   

The parameters for $A_n$'s are fine-tuned by minimizing the following total layer-wise losses for task $\gT_{n}$: 
\begin{equation}
\mathcal{L}_{n}= \sum_{l=1}^{L} {\mathcal{L}}_n^l.
\end{equation}



\section{Experiments}
\subsection{Datasets}
Following the experimental setup of ACM~\citep{zhang2022continual}, we conduct experiments on two scenarios to demonstrate the merits of SAFM: the similar scenario and the dissimilar scenario.  Each scenario contains four task orders as detailed in Appendix~\ref{appendix:orders}.  In the similar scenario, tasks share the same task pattern but originate from different domains. Specifically, we utilize five datasets spanning fourteen domains: E2ENLG~\citep{E2ENLG}, RNNLG~\citep{RNNLG}, Schema Guided Dialogue (SGD)~\citep{rastogi2020towards}, Task-Master 2019 (TM19)~\citep{byrne2019taskmaster}, and Task-Master 2020 (TM20)~\citep{byrne2019taskmaster}.  In the dissimilar scenario, tasks have different task patterns, and the data distribution shifts are substantial.  We apply seven datasets, covering fourteen domains: E2ENLG~\citep{E2ENLG}, RNNLG~\citep{RNNLG}, WikiSQL~\citep{WikiSQL}, CNN/DailyMail~\citep{cnndaily}, SGD~\citep{rastogi2020towards}, TM19~\citep{byrne2019taskmaster}, and TM20~\citep{byrne2019taskmaster}.  The task description and the dataset statistics are reported in Appendix~\ref{appendix:task_description} and Appendix~\ref{appendix:dataset}, respectively.

\subsection{Baselines}
\label{section:bsl}
We evaluate SAFM against strong baselines:
\begin{compactenum}
    \item \textbf{Finetune}~\citep{finetune_bsl} directly fine-tunes the language model on new tasks sequentially. 
    \item \textbf{EWC}~\citep{kirkpatrick2017overcoming} introduces regulation constraints on the loss to prevent updates to crucial parameters from previous tasks.
    \item \textbf{LAMOL}~\citep{sun2019lamol} is a generative replay method that applies a language model as a generator to produce pseudo-samples, training the new task alongside these pseudo-samples to mitigate CF.
    \item \textbf{InsCL}~\citep{wang-etal-2024-inscl} is a strong rehearsal method that dynamically replays previous data based on task similarity using Wasserstein Distance and prioritizes high-quality data through the Instruction Information metric (InsInfo) to assess the instruction complexity and diversity.
    \item \textbf{AdapterCL}~\citep{madotto2021continual}, a robust architectural-based approach, isolates task-specific parameters by creating a dedicated adapter for each task.
    \item \textbf{Adapter+LAMOL}~\citep{zhang2022continual} combines adapters with pseudo-replay generated by a language model via adding a new adapter to each task to learn all tasks sequentially.
    \item \textbf{ACM}~\citep{zhang2022continual} modifies AdapterCL by adaptively reusing previous adapter modules for new tasks, striking a balance between avoiding CF and promoting knowledge sharing.
    \item \textbf{O-LoRA}~\citep{wang-etal-2023-orthogonal} is a parameter-efficient architectural method that learns tasks in different low-rank vector subspaces, which are kept orthogonal to each other to reduce CF.
    \item \textbf{SAPT}~\citep{zhao2024sapt} introduces an adapter for each new task and employs a shared attention framework to enhance knowledge transfer across tasks.
    \item (\textbf{Multi}) performs multi-task learning across all tasks and serves as the upper bound for continual learning performance.
\end{compactenum}

\subsection{Implementation Details}\label{sec:implementation}
The experiments are conducted on an NVIDIA H800-80G GPU. The training time for SAFM is approximately 8 hours in the similar scenario and 14 hours in the dissimilar scenario.
GPT-2~\citep{radford2019language} serves as the backbone language model.
The AdamW optimizer~\citep{loshchilov2017decoupled} is utilized with a learning rate of 1.75e-4.
The batch size is 8. 
During the decision stage, the training epoch is 6, with the initial 3 epochs dedicated to the warmup procedure and the subsequent 3 epochs for the architecture search procedure. 
The sparse factor $\alpha$ is 0.11, and the reuse factor $\beta$ is 0.08.  To prevent the model from getting stuck in a local optimum, such as yielding all empty adapters, we apply no architecture search procedure at layers 5 and 6 for the similar and dissimilar scenarios, respectively.
In the tuning stage, the number of epochs is set to 12. The weight of layer-wise loss is 0.4 for the similar scenario and 0.1 for the dissimilar scenario. 
Following the setup in~\citep{sun2019lamol}, pseudo-replay is implemented with a rate of 0.2.

\renewcommand{\arraystretch}{1.2}
\begin{table*}[!t]
\footnotesize
\resizebox{2\columnwidth}{!}{/
\begin{tabular}{lcccccc}
\hline 
\multirow{3}{*}{\textbf{Methods}} & \multicolumn{3}{c}{\textbf{Similar}} &\multicolumn{3}{c}{\textbf{Dissimilar}} \\
\cline{2-7}
&\multirow{2}{*}{Learn. Param.~$\downarrow$} &\multirow{2}{*}{Score~(\%)~$\uparrow$} & \multirow{2}{*}{BWT~(\%)~$\uparrow$} &\multirow{2}{*}{Learn. Param.~$\downarrow$} &\multirow{2}{*}{Score~(\%)~$\uparrow$} & \multirow{2}{*}{BWT~(\%)~$\uparrow$} \\
&{} &{} & {} &{} & &{}\\
\hline 
Finetune~\citep{finetune_bsl}&1742.30M &15.71 $\pm$ 3.84 &-33.35 $\pm$ 4.24 &1742.30M &7.35 $\pm$ 4.14 &-47.84 $\pm$ 4.61 \\
\hdashline
EWC~\citep{kirkpatrick2017overcoming}&1742.30M &18.23 $\pm$ 4.20 &-30.04 $\pm$ 4.74 &1742.30M &11.35 $\pm$ 5.59 &-43.63 $\pm$ 5.98 \\
\hdashline
LAMOL~\citep{sun2019lamol}&1742.30M &38.40 $\pm$ 2.40 &-8.09 $\pm$ 3.24 &1742.30M &45.81 $\pm$ 3.74 &-6.70 $\pm$ 4.04 \\
InsCL~\citep{wang-etal-2024-inscl}&10780.00M & 41.99 $\pm$ 1.46 &-2.38 $\pm$ 1.73 &10780.00M &47.52 $\pm$ 1.01 &-4.83 $\pm$ 1.01 \\
\hdashline
AdapterCL~\citep{madotto2021continual}&25.06M &44.03 $\pm$ 0.00 &N/A &25.06M &50.82 $\pm$ 0.00 &N/A \\
Adapter+LAMOL~\citep{zhang2022continual} &25.06M &34.39 $\pm$ 1.23 & -11.36 $\pm$ 1.35 &25.06M &44.12 $\pm$ 3.56 &-6.69 $\pm$ 3.99 \\
ACM~\citep{zhang2022continual}&25.06M &41.84 $\pm$ 1.23 &-3.37 $\pm$ 2.09 &25.06M &49.07 $\pm$ 1.53 &-2.29 $\pm$ 1.46 \\
O-LoRA~\citep{wang-etal-2023-orthogonal}&33.03M & 25.93 $\pm$ 1.40 & -17.00 $\pm$ 1.37 & 33.03M & 10.51 $\pm$ 8.47 & -32.71 $\pm$ 7.64 \\
SAPT~\citep{zhao2024sapt}& 55.36M & 42.52 $\pm$ 0.51 & -0.52$\pm$ 0.37 & 55.36M & 40.09$\pm$1.35 & -3.73$\pm$1.31 \\ 
\hdashline
SAFM &\textbf{14.88}M &\textbf{44.55} $\pm$ \textbf{0.30} &\textbf{0.60} $\pm$ \textbf{0.84} &\textbf{15.21}M &\textbf{51.38} $\pm$ \textbf{0.12} &\textbf{0.57} $\pm$ \textbf{0.14} \\
\hline 
Multi (Upper Bound)~\citep{caruana1997multitask} &- & 47.69 & N/A &- & 54.19 & N/A \\
\hline 
\end{tabular}%
}
\caption{Comparison results of SAFM and strong baselines.  The best results are highlighted in bold.  Learn. Param. denotes the number of learnable parameters.  The vertical arrow indicates the direction of a superior model.}
\label{tab:overall}
\end{table*}

\subsection{Evaluation Metrics}
{
The evaluation metric for each task is as follows: 
INTENT uses Accuracy (ACC), evaluating the accuracy between the predicted intent and the real intent.  DST utilizes Joint Goal Accuracy (JGA) ~\citep{wu2019transferable}, where both the intent keyword and its corresponding value must match exactly with the golden truth. 
NLG and summarization employ the BLEU score~\citep{papineni2002bleu}, which measures the similarity between the generated text and the real text. 
SQL Query Generation uses Exact Match (EM), where the generated SQL query should match exactly with the gold.
 }

Additionally, to obtain the overall comparison, we  follow~\citep{madotto2021continual, lopez2017gradient, zhang2022continual} to employ the following two average evaluation metrics: (1) \textbf{Average Score (Score)}~\citep{madotto2021continual, lopez2017gradient, zhang2022continual} defines the average accuracy across all tasks after all the tasks finished learning:
$\mbox{Score} = \frac{1}{t} \sum_{i=1}^{t} R_{N, i}$, where $R_{i, j}$ defines the testing result on task $\gT_{j}$ using the model trained after task $\gT_{i}$.
(2) \textbf{Backward Transfer (BWT)}~\citep{lopez2017gradient, zhu2022continual} quantifies the effects on model performance after training on new tasks, defined by $\mbox{BWT} = \frac{1}{t-1} \sum_{i=1}^{t-1} ( R_{N, i} - R_{i, i} )$.  Both metrics with high values indicate better performance.


\section{Results and Analysis}
\subsection{Main Results} 
Table~\ref{tab:overall} shows the overall performance of SAFM compared to all strong baselines, showcasing improvement in both performance and parameter efficiency across both scenarios:
\begin{compactitem}[\textbullet]
    \item {\bf Superior Performance:} AdapterCL achieves the best performance among all baselines.      SAFM enhances AdapterCL across all metrics.  In the similar scenario, SAFM realizes a 0.52-point increase in the average score while in the dissimilar scenario, it sees a 0.56-point increase.  Significant improvements are also observed in SAPT, the latest competitive architecture-based approach. These gains are attributed to the effective use of adapter fusion and the layer-wise loss mechanism.  The fusion technique improves the learning of global information across tasks, while the layer-wise loss mechanism enhances local information by optimizing parameters, making each adapter more task-specific and increasing the distance between modules within each task.
    
    \item {\bf Positive BWT:} SAFM is the sole method to achieve positive BWT, with values of 0.60 and 0.57 in the similar and dissimilar scenarios, respectively.  A higher BWT indicates better knowledge transfer from new tasks to the previous ones.  The positive BWT reflects SAFM's ability to share knowledge across tasks effectively, thereby alleviating forgetting.

    \item {\bf Parameter Efficiency:} SAFM shows significant parameter reduction compared to baselines such as AdapterCL, and Adapter+LAMOL, and ACM.  This reduction becomes particularly important as the number of tasks grows.  SAFM achieves superior performance using only less than 60\% of the learnable parameters, demonstrating its ability to eliminate redundant adapter layers and optimize model performance with fewer parameters.
 
\end{compactitem}

\begin{figure}[!t]
  \centering
  \includegraphics[width=1\linewidth]{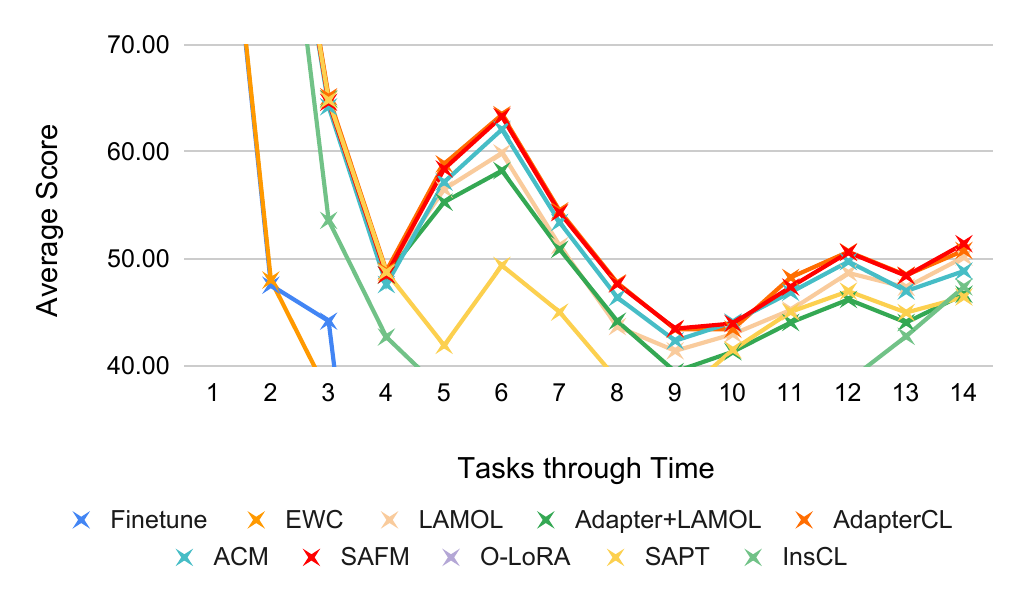}
  \caption{Learning curve of compared methods.  SAFM's position above the other lines indicates its superior performance.
  }
  \label{fig:dis_curve}
  \vspace{-0.8em}
\end{figure}

\renewcommand{\arraystretch}{1.2}
\begin{table*}[htbp]
\scriptsize
\centering
\resizebox{2\columnwidth}{!}{
\begin{tabular}{lccccccccccc}
\hline 
& &\multicolumn{5}{c}{\textbf{Similar}} &\multicolumn{5}{c}{\textbf{Dissimilar}} \\
\cmidrule{3-7} \cmidrule{8-12}
& &Order 1& Order 2 &Order 3 &Order 4 & Avg. &Order 5 &Order 6 &Order 7 &Order 8 & Avg. \\
\hline 
\multirow{3}{*}{ACM} &Score &41.51 &40.58 &43.52 &41.75 &41.84 &48.90 &50.10 &50.67 &47.63 &49.07 \\
&BWT &-4.17 &-5.31 &-0.43 &-3.58 &-3.37 &-2.71 &-1.53 &-0.80 &-4.13 &-2.29 \\
&Learn. Param. &25.06M &25.06M &25.06M &25.06M &25.06M &25.06M &25.06M &25.06M &25.06M &25.06M \\
\hline 
\multirow{3}{*}{SAFM (w/o layer-wise)} &Score &43.44 &43.16 &44.32 &42.92 &43.49 &51.14 &50.95 &50.86 &50.46 &50.85 \\
&BWT &-0.69 &-0.94 &1.12 &-1.69 &-0.55 &0.12 &0.00 &-0.01 &-0.03 &0.02 \\
&Learn. Param. &15.96M &16.56M &14.91M &16.70M &16.06M &15.22M &14.46M &17.00M &17.60M &16.07M \\
\hline 
\multirow{3}{*}{SAFM} &Score &\textbf{44.22} &\textbf{44.56} &\textbf{44.95} &\textbf{44.45} &\textbf{44.55} &\textbf{51.45} &\textbf{51.5} &\textbf{51.45} &\textbf{51.24} &\textbf{51.38} \\
&BWT &\textbf{0.19} &\textbf{0.47} &\textbf{1.82} &\textbf{-0.07} &\textbf{0.60} &\textbf{0.70} &\textbf{0.69} &\textbf{0.46} &\textbf{0.44} &\textbf{0.57} \\
&Learn. Param. &\textbf{15.07M} &\textbf{15.96M} &\textbf{13.87M} &\textbf{14.61M} &\textbf{14.88M} &\textbf{14.91M} &\textbf{13.13M} &\textbf{15.81M} &\textbf{17.00M} &\textbf{15.21M} \\
\hline
\end{tabular}
}
\caption{Ablation study on the impact of architecture search procedure and the layer-wise loss mechanism.  The metrics of `Score', `BWT', and `Learn. Param.' are consistent in Table~\ref{tab:overall}. 
} 

\label{tab:wo}
\end{table*}

\subsection{Ablation Study}
We conduct ablation studies to evaluate the impact of SAFM.  ACM is chosen as the baseline because SAFM is an improved version built upon it.
Additionally, we tested different sizes of GPT-2 as a new backbone to assess the generalization of SAFM.

\paragraph{Learning Curves of Compared Methods}
{
Figure~\ref{fig:dis_curve} presents the learning curve across tasks in Order 5 from Table~\ref{tab:dataset_nask_order} for all compared methods.  To enhance visibility, the figure only includes average scores between 40.0 and 70.0.  Finetune and EWC are omitted after task 3 due to their pronounced susceptibility to CF, highlighting the importance of mitigating forgetting.  Notably, SAFM consistently outperforms other methods, as indicated by its position above all other lines. 
}

\renewcommand{\arraystretch}{1.2}
\begin{table}[!t]
\scriptsize
\centering
{
\centering
\resizebox{0.65\linewidth}{!}{
\begin{tabular}{lccc}
\hline
\multirow{2}{*}{\textbf{Ratio}}  & \multirow{2}{*}{\textbf{Method}} &Order 1 &Order 5 \\
\cmidrule{3-4}
{} & & Score & Score \\\hline
0.2 &ACM  &41.51 &48.90 \\
\hdashline
0.1 &SAFM &42.71 &51.43 \\
0.2 &SAFM &44.22 &51.45 \\
0.5 &SAFM &44.96 &51.55 \\
0.8 &SAFM &45.34 &51.63 \\
\hline
\end{tabular}%
}
}
\caption{Comparison results of ACM with the 0.2 pseudo-sample ratio and SAFM with varying pseudo-sample ratios on both scenarios.}
\label{tab:lamol}
\end{table}

\paragraph{Effect of Key Components in SAFM.}  
Table~\ref{tab:wo} reports the effect of key components in SAFM, specifically the architecture search procedure and the layer-wise Loss mechanism.  The results demonstrate that: (1) Comparing ACM with SAFM without the layer-wise loss, i.e., SAFM (w/o layer-wise), SAFM consistently outperforms ACM across all task orders with a 99\% confidence level in the paired $t$-test, while using only 57.7\%-70.2\% of ACM's learnable parameters.  This suggests that the architecture search procedure indeed reduces parameter redundancy and computational costs by selectively introducing empty or reused adapters, while still absorbing task-specific knowledge more effectively.  (2) After including the layer-wise loss, SAFM further outperforms SAFM (w/o layer-wise) across all task orders, again with a 99\% confidence level on the paired $t$-test, while using fewer parameters.  The improvement underscores the positive impact of the layer-wise loss in enhancing knowledge transfer within the constrained parameter space.

\paragraph{Effect of the Number of Pseudo-Samples.}
To evaluate the impact of the number of pseudo-sample, we conduct experiments using different pseudo-sample ratios in SAFM.  Table~\ref{tab:lamol} compares the effect of the number of pseudo-samples tested in Order 1 and Order 5 from Table~\ref{tab:dataset_nask_order}, representing typical cases of the similar and dissimilar scenarios, respectively.  {For ACM, the ratio of pseudo-samples is fixed at 0.2, meaning that the number of generated pseudo-samples equals 20\% of the training data in the current task.}  For SAFM, the ratio varies from \{0.1, 0.2, 0.5, 0.8\}.  The results show that: (1) SAFM with only a 0.1 pseudo-sample ratio outperforms ACM with a 0.2 ratio, which demonstrates SAFM's parameter efficiency and memory saving, absorbing more knowledge with fewer pseudo-samples.  (2) As the pseudo-sample ratio increases, SAFM consistently improves across both test cases, although the improvement in the assimilate scenario is gradual.  This trend is expected, as SAFM requires more memory to assimilate knowledge from previous tasks.

\renewcommand{\arraystretch}{1.2}
\begin{table}[ht]
\scriptsize
\centering
{
\resizebox{\linewidth}{!}{
\begin{tabular}{c|cc|c|cc}
\hline
\multirow{2}{*}{\textbf{Layer Index}} &\multicolumn{2}{c|}{Order 1} & \multirow{2}{*}{\textbf{Layer Index}} &\multicolumn{2}{c}{Order 5} \\
{} &ACM &SAFM & {} &ACM &SAFM \\
\hline
Null&41.51 &42.58 &Null&48.90 &50.95 \\
3 &42.27 &43.44 &2 &50.62 &51.02 \\
5 &42.98 &44.22 &4 &50.76 &51.35 \\
7 &43.64 &44.26 &6 &50.94 &51.45 \\
9 &44.12 &44.31 &8 &51.27 &51.56 \\
\hline
\end{tabular}%
}
}
\caption{{Score of no AS adapter layer in Order 1 and Order 5.  `Null' indicates no AS applied in all layers.}}
\label{tab:layer_idx}
\vspace{-1em}
\end{table}

{
\paragraph{Effect of No Architecture Search (AS) in Adapter Layers.}
We evaluate the effect of conducting no AS in an adapter layer of SAFM, i.e., specifically assigning a layer without performing the AS procedure.  The selected layer varies from \{3, 5, 7, 9\} in Order 1 for the similar scenario and \{2, 4, 6, 8\} in Order 5 for the dissimilar scenario, respectively, where `Null' denotes no AS applied in all layers.  Table~\ref{tab:layer_idx} shows that: (1) Integrating a no AS adapter layer enables the exclusive retention of task-specific knowledge, enhancing performance.  This approach preserves more high-level task-specific information, leading to improved outcomes.  Note that, via the setting detailed in Sec.~\ref{sec:implementation}, the results of applying no AS to layer 5 in Order 1 and layer 6 in Order 5 match the corresponding performance in Table~\ref{tab:wo}.
(2) Both ACM and SAFM demonstrate improved performance as the layer index increases in both cases.  This enhancement is likely due to higher layers containing more meaningful, complex, and high-level information, as noted in~\citep{erhan2009visualizing}. 
}

\renewcommand{\arraystretch}{1}
\begin{table*}[ht]
\footnotesize
\centering
\begin{tabular}{lp{13.5cm}}
\hline
\textbf{E2ENLG} & \textbf{name[Green Man], eatType[pub], customer rating[3 out of 5], near[All Bar One]} \\
\hdashline
Reference & Located close to All Bar One, Green Man pub has a 3 out of 5 rating. \\
ACM & Near All Bar One it has a customer rating of 3 out of 5. \\
SAFM & Green Man is a pub near All Bar One with a customer rating of 3 out of 5. \\
\midrule
\textbf{WikiSQL} & \textbf{The table has columns of [} \\
        & \textbf{\qquad"Rank Each wrestlers total number of days as champion are ranked highest to lowest;} \\
        & \textbf{\qquad wrestlers with the same number mean that they are tied for that certain rank."}, \\
        & \textbf{\qquad"Wrestler", "\# of reigns", "Combined defenses", "Combined days"} \\
        & \textbf{\quad]}, \\
        & \textbf{Question: "In terms of reigns, what is the lowest number listed?"} \\
\hdashline
Reference & SELECT MIN \# of reigns FROM table \\
ACM & SELECT MIN number of days as champion FROM table WHERE rank = highest to lowest \\
SAFM & SELECT MIN \# of reigns FROM table \\
\bottomrule
\end{tabular}
\caption{Comparison of the Ground Truth (Reference) with the generated outputs from ACM and SAFM.}
\label{tab:casestudy}
\vspace{-1em}
\end{table*}

\paragraph{Scale-up of different backbones.}
To evaluate the scalability of SAFM, we conducted experiments using GPT-2 of different sizes and Llama3-8B \cite{touvron2023llama}. Table \ref{tab:scaleup} reports the average performance of SAFM and ACM after training on task 8 and task 10 in Order 1.
The results show that SAFM consistently outperforms ACM on GPT-2 of different sizes and Llama3-8B. This aligns with the scaling law \cite{kaplan2020scaling,hoffmann2022training}, which highlights the consistency of pre-trained decoder-only models. SAFM is proven to significantly improve performance over ACM, proving its scalability and generalizability in different backbone language models (LMs).  However, the performance of Llama3-8B is significantly lower compared to fine-tuning with GPT-2, primarily due to the limited availability of training data. The conclusion is aligned with DCL \cite{zengdirichlet}.

\renewcommand{\arraystretch}{1.5}
\begin{table}[ht]
  \centering
  \resizebox{\linewidth}{!}{
  \begin{tabular}{lcccc}
        \hline
        Backbone             &ACM  & Learn. Param. & SAFM & Learn. Param.\\
     
        \hline             
                                & \multicolumn{4}{c}{Task 8 }  \\ 
                                \cline{2-5}
        GPT-2 (124M)                     & 56.11 & 14.3M & \textbf{56.55} &6.7M\\
        GPT-2-medium (355M)              & 48.36 & 50.7M& \textbf{50.08} & 34.0M\\
        GPT-2-large (774M)              & 48.05 &118.7M & \textbf{49.53} &83.6M\\
        Llama3-8B              & 20.95 &5279.3M & \textbf{42.78} &1031.1M \\
    \hline
                                & \multicolumn{4}{c}{Task 10 }  \\ 
                                \cline{2-5}
        GPT-2 (124M)                     & 49.35 &17.9M & \textbf{49.39} & 8.7M\\
        GPT-2-medium (355M)              & 40.77 & 63.4M& \textbf{43.36} &38.8M\\
        GPT-2-large (774M)              & 39.33 &148.4M & \textbf{42.94} &130.6M\\
         Llama3-8B              & 16.23 &6599.1M & \textbf{36.44} &1299.1M\\
    \hline
  \end{tabular}
   }
  \caption{
  Scores of SAFM and ACM with different backbones in Order 1.}
  \label{tab:scaleup}
\end{table}

{
\subsection{Case Study}
Table~\ref{tab:casestudy} presents a comparison of the generated samples of SAFM, ACM, and the ground truth (Reference).  We selected the E2ENLG dataset used in Order 1 and the WikiSQL dataset used in Order 8 as two representative samples to illustrate the output generated by different methods. 
In the E2ENLG dataset, the example \texttt{``name[Green Man], eatType[pub], customer rating[3 out of 5], near[All Bar One]''} illustrates that ACM overlooks two essential pieces of information in the provided structured data:  \texttt{``name[Green Man]''} and \texttt{``eatType[pub]''}.
In contrast, SAFM successfully captures all critical information, showing its superior capability in enhancing model performance.

In the WikiSQL dataset examples, ACM faces challenges in understanding the structured data, distinguishing between natural language and SQL, and interpreting the question intent.  As illustrated in Table~\ref{tab:casestudy}, ACM misinterprets the column \texttt{``\# of reigns''} as the natural language phrase \texttt{``number of days as champion''}.  Additionally, ACM generates SQL content related to \texttt{``rank''}, even though there is no connection to \texttt{``rank''} in the provided question.  Consequently, ACM fails to generate the desired SQL statement accurately.  In contrast, SAFM's generation aligns precisely with the ground truth, showcasing its superior performance.
}

\section{Conclusion}
This paper presents SAFM, a novel approach that efficiently leverages both global and local information to address catastrophic forgetting in continual learning.  SAFM operates through a two-stage process in which adapter layers are strategically abandoned, reused, or added during the decision stage to facilitate knowledge sharing and reduce parameter usage.  Additionally, a layer-wise loss mechanism is introduced in the tuning stage to optimize knowledge representation within the limited parameter budget.  Our extensive analysis of SAFM examines the impact of the architecture search procedure, the layer-wise loss, the percentage of pseudo-replay, and the effects of various fixed adapter layers.  Experimental results consistently demonstrate that SAFM outperforms SOTA methods by utilizing only less than 60\% of the parameters employed by the SOTA models.


\section{Limitations}
SAFM primarily focuses on task-level incremental information and does not explicitly address sample-level incremental information, which may result in reduced performance on outlier samples.

\bibliography{anthology,custom}
\newpage
\appendix
\section{Task Description}
\label{appendix:task_description}
\noindent
\textbf{INTENT} aims to categorize user queries into specific intents. 
For example, intent for \texttt{``USER: I want to book two tickets for Star Wars.''} is \texttt{``movie\_booking.''}, while intent for \texttt{``USER: I need to check my balance.''} is \texttt{``CheckBalance''}.\\
\textbf{DST} involves tracking the state of dialogue, such as the topic discussed, the emotion expressed, or any other relevant data points. 
For example, in a flight booking conversation, \texttt{``USER: I want to fly to Seattle this evening.''}.  
The user's intent \texttt{``Destination''} can be identified from \texttt{``fly to''}, and the corresponding value is \texttt{``Seattle''}. \\
\textbf{NLG} generates human-like text from structured data. For example, given output from DST \texttt{``Destination: Seattle, Price: 200 dollars, Departure Time: 7.50 PM''}, the generation of the system might be \texttt{``Dear sir, the ticket price for the flight to Seattle on 7.50 PM is 200 dollars''}.\\
\textbf{Summarization} aims at condensing a long text into a shorter one without missing important information and overall meaning.\\
\textbf{SQL Query Generation} refers to automatically generating SQL queries, which are used to retrieve or manipulate data from relational databases. 
It involves transforming high-level user requirements and structured data into executable SQL statements.
For example, the user requirement is \texttt{``select the max price on date '2024-06-04' in given sheet''}, and the SQL query should be \texttt{``SELECT MAX price  FROM sheet WHERE date = '2024-06-04';''}.\\

\renewcommand{\arraystretch}{1.2}
\begin{table*}[!t]
\small
\centering
\resizebox{2\columnwidth}{!}{
\begin{tabular}{lllllll}
\hline
Name &  Domain & Dataset & Task Pattern & Train  & Valid & Test  \\
\hline
CNN/DailyMail & news & CNN/DailyMail & Summarization & 6604 & 2250 & 2250 \\
WikiSQL & programming language & WikiSQL & \footnotesize{SQL Query Generation} & 6525 & 8421 & 15878\\
E2ENLG & restaurant & E2ENLG & NLG & 6000 & 2000 & 2000 \\
RNNLG\_hotel & hotel & RNNLG & NLG & 6446 & 1075& 1075 \\
RNNLG\_rest & restaurant & RNNLG & NLG & 6228 & 1039 & 1039 \\
RNNLG\_tv & tv & RNNLG & NLG & 8442 & 1407 & 1407 \\
RNNLG\_laptop & laptop & RNNLG & NLG & 7944 & 2649 & 2649 \\
SGD\_hotel\_nlg & hotel & SGD & NLG & 1997 & 243 & 597 \\
SGD\_restaurant\_nlg & restaurant & SGD & NLG & 1720 & 166 & 386 \\
SGD\_restaurant\_intent & restaurant & SGD & INTENT & 2686 & 278 & 616 \\
SGD\_restaurant\_dst & restaurant & SGD & DST & 2686 & 278 & 616 \\
SGD\_flight\_nlg  & flight & SGD & NLG & 2571  & 627 & 982 \\
SGD\_flight\_intent & flight & SGD & INTENT & 4766 & 1041 & 1756 \\
SGD\_flight\_dst & flight & SGD & DST & 4766 & 1041 & 1756 \\
TM19\_restaurant\_nlg  & restaurant & TM19 & NLG & 2582 & 330 & 333 \\
TM19\_movie\_nlg  & movie & TM19 & NLG  & 3010 & 366 & 341 \\
TM20\_hotel\_nlg & hotel & TM20 & NLG & 6590 & 842 & 869 \\
TM20\_restaurant\_nlg & restaurant & TM20 & NLG & 8356 & 1063 & 994 \\
TM20\_restaurant\_intent & restaurant & TM20 & INTENT & 13738 & 1761 & 1791 \\
TM20\_restaurant\_dst & restaurant & TM20 & DST & 13738 & 1761 & 1791\\
TM20\_flight\_nlg & flight & TM20 & NLG & 10148 & 1272 & 1245\\
TM20\_flight\_intent & flight & TM20 & INTENT & 15868 & 1974 & 1940 \\
TM20\_flight\_dst & flight & TM20 & DST & 15868 & 1974 & 1940 \\
TM20\_movie\_nlg & movie & TM20 & NLG & 9406 & 1203 & 1093 \\
\bottomrule
\end{tabular}}
\caption{Dataset Statistics.}
\label{tab:dataset_sta}
\end{table*}

\section{Dataset Statistics}
\label{appendix:dataset}

We describe the details of the datasets as follows:
\begin{compactitem}[\textbullet]
    \item \textbf{E2ENLG}~\citep{E2ENLG} is a dataset that focuses on NLG data in the restaurant domain.
    \item \textbf{RNNLG}~\citep{RNNLG} is a dataset that includes NLG data for spoken dialogue systems. It covers four domains: restaurant, TV, laptop, and hotel. 
    \item     \textbf{SGD}~\citep{rastogi2020towards} is a dataset that contains multi-domain, task-oriented conversations between a user and a virtual assistant.  It reflects real-world scenarios by including different APIs with overlapping functionalities but different interfaces. 
    \item \textbf{TM19}~\citep{byrne2019taskmaster} is a task-based dataset comprising spoken and written dialogues created through two distinct procedures.  It covers six domains: ordering pizza, creating auto repair appointments, setting up ride service, ordering movie tickets, ordering coffee drinks, and making restaurant reservations.
    \item \textbf{TM20}~\citep{byrne2019taskmaster} is a dataset that includes dialogues in seven domains: restaurants, food ordering, movies, hotels, flights, music, and sports.  It consists entirely of spoken two-person dialogues and contains many search-oriented and recommendation-oriented conversations. 
    \item \textbf{CNN/DailyMail}~\citep{cnndaily} is a dataset that contains news stories from CNN and Daily Mail, along with human-generated abstractive summaries. 
    \item \textbf{WikiSQL}~\citep{WikiSQL} comprises hand-annotated examples of questions and SQL queries.
\end{compactitem}
Table~\ref{tab:dataset_sta} provides a detailed summary of the dataset statistics.

\section{Task Orders}
\label{appendix:orders}
Table \ref{tab:dataset_nask_order} describes the details of the task orders. 

\renewcommand{\arraystretch}{1.2}
\begin{table*}[!t]
\small
\centering
\resizebox{2\columnwidth}{!}{
\begin{tabular}{ccl}
\hline
\textbf{Order} & \textbf{Scenario}  & \textbf{\qquad \qquad \qquad \qquad \qquad \qquad Task Sequence} \\
\hline

\multirow{3}{*}{1} & \multirow{3}{*}{Similar}  & {E2ENLG $\rightarrow$ RNNLG\_rest $\rightarrow$ RNNLG\_hotel $\rightarrow$ SGD\_hotel\_nlg $\rightarrow$ RNNLG\_laptop $\rightarrow$} \\
{} & {} & {RNNLG\_tv $\rightarrow$ SGD\_restaurant\_nlg $\rightarrow$ SGD\_flight\_nlg $\rightarrow$ TM19\_movie\_nlg $\rightarrow$ TM20\_hotel\_nlg $\rightarrow$} \\
 {} & {} & {TM20\_restaurant\_nlg $\rightarrow$ TM20\_flight\_nlg $\rightarrow$ TM20\_movie\_nlg $\rightarrow$ TM19\_restaurant\_nlg} \\
 \midrule
 
 \multirow{3}{*}{2} & \multirow{3}{*}{Similar}  & {RNNLG\_hotel $\rightarrow$ SGD\_hotel\_nlg $\rightarrow$ E2ENLG $\rightarrow$ RNNLG\_rest $\rightarrow$ RNNLG\_laptop $\rightarrow$} \\
{} & {} & {SGD\_flight\_nlg $\rightarrow$ SGD\_restaurant\_nlg $\rightarrow$ RNNLG\_tv $\rightarrow$ TM19\_movie\_nlg $\rightarrow$ TM20\_hotel\_nlg $\rightarrow$} \\
 {} & {} & {TM20\_flight\_nlg $\rightarrow$TM20\_movie\_nlg $\rightarrow$TM20\_restaurant\_nlg $\rightarrow$ TM19\_restaurant\_nlg} \\
 \midrule

\multirow{3}{*}{3} & \multirow{3}{*}{Similar}  & {SGD\_flight\_nlg $\rightarrow$ SGD\_restaurant\_nlg $\rightarrow$ E2ENLG $\rightarrow$ RNNLG\_rest $\rightarrow$ RNNLG\_tv $\rightarrow$} \\
{} & {} & {RNNLG\_laptop $\rightarrow$ TM19\_movie\_nlg $\rightarrow$ RNNLG\_hotel $\rightarrow$ SGD\_hotel\_nlg $\rightarrow$ TM19\_restaurant\_nlg $\rightarrow$} \\
 {} & {} & {TM20\_flight\_nlg $\rightarrow$ TM20\_hotel\_nlg $\rightarrow$ TM20\_restaurant\_nlg $\rightarrow$ TM20\_movie\_nlg} \\
 \midrule

\multirow{3}{*}{4} & \multirow{3}{*}{Similar}  & {E2ENLG $\rightarrow$ RNNLG\_rest $\rightarrow$ RNNLG\_hotel $\rightarrow$ RNNLG\_tv $\rightarrow$ RNNLG\_laptop $\rightarrow$} \\
{} & {} & {SGD\_hotel\_nlg $\rightarrow$ SGD\_flight\_nlg $\rightarrow$ SGD\_restaurant\_nlg $\rightarrow$ TM19\_movie\_nlg $\rightarrow$ TM19\_restaurant\_nlg $\rightarrow$} \\
 {} & {} & {TM20\_hotel\_nlg $\rightarrow$ TM20\_movie\_nlg $\rightarrow$ TM20\_flight\_nlg $\rightarrow$ TM20\_restaurant\_nlg} \\
\midrule

\multirow{3}{*}{5} & \multirow{3}{*}{Dissimilar}  & {SGD\_restaurant\_intent $\rightarrow$ SGD\_flight\_intent $\rightarrow$ SGD\_restaurant\_dst $\rightarrow$ SGD\_flight\_dst $\rightarrow$ TM20\_restaurant\_intent $\rightarrow$  } \\
{} & {} & {TM20\_flight\_intent $\rightarrow$ TM20\_restaurant\_dst $\rightarrow$ TM20\_flight\_dst $\rightarrow$ CNN/DailyMail $\rightarrow$ WikiSQL $\rightarrow$ } \\
 {} & {} & {RNNLG\_laptop $\rightarrow$ RNNLG\_tv $\rightarrow$ E2ENLG $\rightarrow$ RNNLG\_hotel} \\
\midrule

\multirow{3}{*}{6} & \multirow{3}{*}{Dissimilar}  & {SGD\_restaurant\_intent $\rightarrow$ SGD\_flight\_intent $\rightarrow$ TM20\_restaurant\_intent $\rightarrow$ TM20\_flight\_intent $\rightarrow$ E2ENLG $\rightarrow$ } \\
{} & {} & {RNNLG\_hotel $\rightarrow$ RNNLG\_laptop $\rightarrow$ RNNLG\_tv $\rightarrow$ SGD\_restaurant\_dst $\rightarrow$ SGD\_flight\_dst $\rightarrow$ } \\
 {} & {} & {TM20\_restaurant\_dst $\rightarrow$ TM20\_flight\_dst $\rightarrow$ CNN/DailyMail $\rightarrow$ WikiSQL} \\
\midrule

\multirow{3}{*}{7} & \multirow{3}{*}{Dissimilar}  & {TM20\_restaurant\_intent $\rightarrow$ TM20\_flight\_intent $\rightarrow$ SGD\_restaurant\_intent $\rightarrow$ SGD\_flight\_intent $\rightarrow$ E2ENLG $\rightarrow$  } \\
{} & {} & {RNNLG\_hotel $\rightarrow$  CNN/DailyMail $\rightarrow$  WikiSQL $\rightarrow$  RNNLG\_laptop $\rightarrow$  RNNLG\_tv $\rightarrow$ } \\
 {} & {} & {SGD\_restaurant\_dst $\rightarrow$  SGD\_flight\_dst $\rightarrow$  TM20\_restaurant\_dst $\rightarrow$  TM20\_flight\_dst} \\
\midrule

\multirow{3}{*}{8} & \multirow{3}{*}{Dissimilar}  & {WikiSQL $\rightarrow$ CNN/DailyMail $\rightarrow$ TM20\_restaurant\_intent $\rightarrow$ SGD\_restaurant\_intent $\rightarrow$ SGD\_flight\_intent $\rightarrow$ } \\
{} & {} & {TM20\_flight\_intent $\rightarrow$ SGD\_restaurant\_dst $\rightarrow$ SGD\_flight\_dst $\rightarrow$ RNNLG\_hotel $\rightarrow$ RNNLG\_laptop $\rightarrow$ } \\
 {} & {} & {E2ENLG $\rightarrow$ TM20\_restaurant\_dst $\rightarrow$  RNNLG\_tv $\rightarrow$ TM20\_flight\_dst } \\

\midrule

\end{tabular}%
}
\caption{Eight random task orders for evaluating SAFM.}
\label{tab:dataset_nask_order}
\end{table*}



\end{document}